%% file: main.tex
\documentclass[11pt]{article}
\usepackage[margin=1in]{geometry}
\usepackage[T1]{fontenc}
\usepackage[utf8]{inputenc}
\usepackage{amsmath}
\usepackage{amssymb}
\usepackage{array}
\usepackage{booktabs}
\usepackage{placeins}
\usepackage{graphicx}
\usepackage[numbers,sort&compress]{natbib}
\usepackage[colorlinks=true,linkcolor=blue,citecolor=blue,urlcolor=blue]{hyperref}
\usepackage{xspace}

\newcommand{\verified}{\textsc{verified}\xspace}
\newcommand{\weak}{\textsc{weak}\xspace}
\newcommand{\unverified}{\textsc{unverified}\xspace}
\newcommand{\contradicted}{\textsc{contradicted}\xspace}
\title{Auditing the Synthetic Memoir: Measuring Scene-Level Confabulation
in LLM-Generated Autobiography Against the Documented Record
of the Life It Describes}
\author{Heather Renze\\
Serenze Global\\
\texttt{serene@serenze.org}\\
ORCID: \href{https://orcid.org/0009-0001-8789-4611}{https://orcid.org/0009-0001-8789-4611}}
\date{August 22, 2026}

\begin{document}
\maketitle

\begin{abstract}
When a large language model (LLM) is asked to write a person's life, how much of what
it writes actually happened? We present a scene-level case-study audit --- the first
quantified audit of LLM-generated autobiography against a subject-specific ground-truth
corpus that we are aware of, based on an unsystematic literature search. The subject
and the author of this paper are the same person: a 366-day ``page-a-day'' book of
first-person anecdotal entries was drafted with a conversational LLM whose documented
inputs were a template, two exemplar days, and each day's quote --- not her corpus ---
and every day was subsequently audited at the anecdote-scene level against an
independent verification corpus using a four-level rubric fixed before analysis.
We define the \emph{verification-failure rate} as the share of days not rated
\verified{} (scene positively corroborated): 354 of 366 days fail, 96.7\%
(Wilson 95\% CI 94.4--98.1\%). Only 12 days contain a corroborated scene; 19 days
(5.2\%) assert claims actively contradicted by the record; the dominant failure mode
is \emph{grounded drift} --- real people, employers, and settings inside invented
scenes --- though its measured share varies across raters. Independent re-rating
replicates the headline (no evidence the original rate was inflated) while showing
that the four-way taxonomy has only fair-to-moderate reliability. Regenerating the
same days with current named models reproduces 100\% verification failure under the
same inputs; grounding generation in the subject's corpus significantly improves the
verification rate while leaving substantial residual failure (83.3\%). We contribute
the measurement, a reusable audit instrument whose WEAK/UNVERIFIED boundary we show
to be unreliable, and a grounding remedy with quantified effect.
\end{abstract}

\section{Introduction}\label{sec:intro}

Systems that simulate people from their data are moving from research prototypes
\citep{park2023generative,park2024selfreports,wang2025beyondprofile} into personal use:
digital twins, synthetic biographies, and ghost-written life narratives built by
prompting an LLM with a person's corpus and asking it to write ``her memories'' in the
first person. Evaluation of such systems has concentrated on attitudes, survey answers,
and surface facts \citep{park2024selfreports,wang2025beyondprofile}. The episodic layer
--- the specific scenes a system asserts about a life --- has not been audited against
the documented record of that life, although laboratory work shows conversational AI can
implant false memories in human subjects \citep{chan2024falsememories}.

This paper reports a complete, naturalistic instance of exactly this failure mode, caught
and measured before publication. Between drafting sessions, a professional author (the
present paper's author) produced a year-long gift book: 366 daily entries ($\approx$543 words each on average; 198{,}949 words in total), written
in the first person about her own life. Repository
documentation states plainly that the entries were ``AI-padded'': drafted with a
conversational LLM (generator documented, Section~\ref{sec:limitations}; its
documented inputs were the template and the day's quote --- see below). Before the derived page-a-day edition
was assembled, every one of the 366 days was fact-checked at the scene level against an
independent ground-truth corpus assembled for that purpose, using a rubric fixed on
July~13, 2026 --- before any analysis in this paper was performed.

We ask three questions. \textbf{RQ1:} What fraction of generated anecdotal ``memories''
survive scene-level verification against an independent record? \textbf{RQ2:} What
\emph{forms} does the failure take? \textbf{RQ3:} Where do contradictions concentrate,
and what remediation does that suggest?

The study is falsifiable in the ordinary way. Had half or more of the 366 scenes been
rated \verified, the framing ``generation without grounding'' would fail; had verification failures
been dominated by \unverified rather than \weak days, the ``grounded drift'' account in
Section~\ref{sec:discussion} would fail. Both outcomes were live possibilities before
tallying.

We contribute: (a)~a quantified audit of confabulation in
personal-narrative generation against subject-specific ground truth ($n=366$
day-narratives; 96.7\% fail positive scene-level verification under our definition); (b)~codification
and release of a reusable scene-level audit instrument --- the four-verdict rubric quoted
verbatim plus a nine-theme recurring-false-premise taxonomy operationalized as fixed,
published keyword screens; and (c)~a remediation workflow (lesson-only rewrite;
author-adjudication precedence) specified during the same project; and (d)~an
inter-rater reliability analysis with two independent re-raters, yielding the
paper's most transferable methodological finding: scene-level fabrication audits are
reliable at the binary level (does this day's scene check out?) but only fair at the
taxonomic level, because the rubric's WEAK/UNVERIFIED boundary turns on an undefined
notion of ``setting.''

A note on positionality shapes everything that follows: the auditing author and the
audited subject are the same person. Section~\ref{sec:ethics} treats the consent,
privacy, and self-audit implications; Section~\ref{sec:single-rater} treats the methodological
ones.

\section{Related Work}\label{sec:related}

\textbf{Hallucination and confabulation in LLMs.} Hallucination is surveyed extensively:
factuality vs.\ faithfulness taxonomies \citep{huang2025survey,ji2023survey}, mitigation
surveys \citep{zhang2023siren,tonmoy2024mitigation}, and a 2026 comprehensive review
\citep{alansari2026comprehensive}. \citet{sui2024confabulation} argue for the term
\emph{confabulation} --- fluent, confident fabrication that fills gaps --- which we adopt:
the audited entries are not random noise but coherent, confident first-person narrative.
ReFACT \citep{wang2025refact} benchmarks scientific confabulation detection with
fine-grained annotations; our instrument is analogous but subject-grounded rather than
encyclopedic.

\textbf{Factuality measurement of long-form generation.} FActScore decomposes long-form
text into atomic claims and checks each against a knowledge source
\citep{min2023factscore}; long-form factuality benchmarks evaluate models against world
facts \citep{wei2024longform}. Our audit is a scene-level analogue: a day-entry is
decomposed into its anecdote scene, and the reference work is not an encyclopedia but an
independent corpus documenting one life. No prior benchmark we are aware of uses a
verified human-life corpus as ground truth.

\textbf{Person simulation and digital twins.} Generative agents produce believable
behavior from profiles and memories \citep{park2023generative}. Interview-grounded agents
simulate 1{,}000 real individuals \citep{park2024selfreports}, with fidelity evaluated on
attitudes and survey responses; persona work explicitly separates surface-fact from deep
persona fidelity \citep{wang2025beyondprofile}; digital-twin construction from
conversation logs is demonstrated for counselors \citep{xie2024psydt}. Our results supply
the missing episodic dimension: even \emph{surface} episodic content --- employers, dates,
named events --- fails verification in the large majority of generated days.

\textbf{False memories and identity risks.} \citet{chan2024falsememories} show LLM
conversational partners amplify false memories in witness interviews --- the
\emph{consumption} side of the risk. Our study documents the complementary
\emph{production} side at scale: an LLM asked to \emph{write} a person's past produces
unsupported scenes in 96.7\% of days under our definition. Persona framing also perturbs factual QA
\citep{akpinar2025whosasking}, consistent with our observation that first-person
narrative framing invites invention.

\textbf{Character consistency failures.} RoleBreak and SHARP treat character
hallucination in role-playing systems as an attack surface
\citep{tang2024rolebreak,kong2024sharp}. We quantify the same drift class when the
``character'' is a real, documented person and the drift is measured against her actual
records rather than a prompt specification.

\textbf{Gap.} Hallucination is surveyed; long-form factuality is benchmarked against
encyclopedic truth; person simulators are built and evaluated on attitudes; false-memory
risks are shown in laboratories; character drift is characterized adversarially.
\textbf{No published study audits LLM-generated autobiography scene-by-scene against the
documented record of the person it describes.} This paper fills that gap.

\section{Data and Study Design}\label{sec:data}

The study stands on three corpora, all fixed before this paper's analysis session
(August~21, 2026); the audit labels themselves predate it by five weeks.

\textbf{Suspect corpus (what was audited).} A year-long ``page-a-day'' book of 366 daily
first-person entries ($\approx$543 words/day; 198{,}949 words in total), drafted for a
gift edition. The source
calendar's own README documents the entries as ``AI-padded'': long-form drafts written
with a conversational LLM. Session-log provenance recovery
(Section~\ref{sec:limitations}) has since documented the generator of record: OpenAI
o3-pro designed the template and gpt-4o / o3 / o4-mini-high generated entries from the
quote list in ChatGPT conversations (July--August 2025), with a Cursor composer run on
August~1, 2025 executing a one-shot prompt. A derived edition
compresses each entry to a $\approx$175-word page (64{,}198 words total). The audit
month-files are headed ``page-a-day,'' but their anecdote rows correspond scene-for-scene
to the \emph{long-form} entries, not the compressed pages: the May~1 audit row, for
example, describes an Evernote scene (three stakeholders' deliverables, crying at 2 a.m.)
that appears only in the long source, and the October~10 row's ``empty Pacific horizon''
phrasing likewise matches only the long text. We therefore treat the long-form entries as
the text that was audited. What the generator received is documented precisely
(Section~\ref{sec:limitations}; \texttt{results/original\_generation\_prompt.txt}):
a template, two exemplar days carrying personal facts, and the day's quote from
\texttt{commonsense.csv} --- \emph{not} the ground-truth corpus. Only replication
Arm~C receives corpus excerpts; the 2025 original and Arms B/B2 receive the
template-plus-quote inputs alone; any additional biographical material the 2025 run
carried in conversation context or drew from pretraining is undocumented. This matters
for interpretation: the ``grounded drift'' pattern of Section~\ref{sec:results} arose
without the record in front of the model. The audited unit is the
\emph{day-entry}, each organized around one anecdotal ``memory'' scene plus a lesson. The
book project's build pipeline (a derived $\approx$150--200-word page per day, plus a
367-row quote index) derives mechanically from these entries; the day-entries are the
atomic object of audit.

\textbf{Ground-truth corpus (what was checked against).} Assembled by the audit to be
independent of generation:
(i)~the memoir manuscript \emph{After the Unicorn} with its editorial anecdote ledgers;
(ii)~the memoir \emph{Birth of a Unicorn};
(iii)~a 353k-word corpus of the subject's published writing and talk transcripts
(HEATHER\_CORPUS);
(iv)~a speaking record and knowledge base maintained for her professional digital twin;
(v)~a music catalogue with dated life-era annotations;
(vi)~targeted web checks.
The calendar source itself, its page-a-day derivative, and the ChatGPT conversation
backups were \emph{excluded} as non-independent --- they could not corroborate what they
generated.

\textbf{Audit labels (the measurements).} Every day received a scene-level verdict on
July~13, 2026, recorded in twelve month tables (one row per day: date, anecdote gist,
verdict, supporting source, evidence note), summarized in a fact-check summary document.
Critical items were adjudicated by the subject directly; four explicit author
confirmations are recorded, and the summary states that author confirmations take
precedence over any ruling. This label set --- not any new judgment --- is the dataset
analyzed here; Sections~\ref{sec:parsing}--\ref{sec:results} are strictly mechanical
parsing and aggregation of documents that existed before this paper was begun.

\textbf{Timeline.} Entries generated July--August 2025 (documented; per-day dates
unrecorded) $\to$
scene-level audit completed July~13, 2026 $\to$ remediation designed (same date) $\to$
dataset parsed and analyzed August~21, 2026. No audit label was created or modified by
this paper.

\section{Method}\label{sec:method}

\subsection{Verdict taxonomy}\label{sec:taxonomy}

The rubric below is quoted verbatim from the audit summary dated July~13, 2026
(\path{FACT_CHECK_SUMMARY.md}); it is a pre-stated instrument, not a post hoc scheme.
Each day's central anecdote scene received exactly one verdict:

\begin{itemize}
\item \verified{} --- ``specific scene corroborated'';
\item \weak{} --- ``real setting/person, invented specific scene'';
\item \unverified{} --- ``no anchor; generic invention'';
\item \contradicted{} / flagged --- ``asserts something FALSE about her life.''
\end{itemize}

The classes are ordered by evidential relation to the record, not by severity alone:
\verified requires positive corroboration; \unverified records absence of evidence;
\contradicted requires positive disproof. \weak is the pivotal class for person
simulation: the system correctly retrieved real grounding material (employer, city,
person) and invented the specifics around it.

\subsection{Audit procedure}\label{sec:procedure}

For each of the 366 days, the auditor extracted the day's anecdote scene from the entry,
searched the ground-truth corpus for corroboration or contradiction, recorded the
verdict with a supporting source citation and an evidence note in the month table, and
flagged candidate false premises. Where a claim turned on facts only the subject could
settle (e.g., whether a childhood accident occurred at age 13 or in adulthood; whether a
treatment history included chemotherapy), the subject issued an explicit written ruling,
and the summary's precedence rule --- ``Author confirmations (2026-07-13, take
precedence)'' --- binds all other rows. Four such rulings are recorded. Health-related
content in this paper is restricted to what those rulings themselves state.

\subsection{Parsing and analysis procedures}\label{sec:parsing}

Two scripts, both published with the paper, convert the audit tables into every number in
Section~\ref{sec:results}.

\texttt{01\_parse\_factcheck.py} reads the twelve month tables and emits one CSV row per
day. Verdict cells are normalized by fixed rules in fixed order: any cell containing
``contradicted'' $\to$ \contradicted; else prefix ``verified'' $\to$ \verified; else
prefix ``unverified'' $\to$ \unverified; else prefix ``weak'' $\to$ \weak; anything else
is reported as unparsed, never silently dropped. Rows are keyed on parsed month/day; the
parser reports duplicates and unparsed candidate lines explicitly. Output:
366 rows, zero duplicate keys, zero unparsed verdicts (\texttt{parse\_report.txt}).

\texttt{02\_analyze.py} asserts $n=366$, tallies verdicts overall and per month, computes
Wilson score intervals ($z=1.96$), applies the nine fixed keyword screens of
Section~\ref{sec:premises} over each row's gist+note text, codes each populated Source
cell against seven named source classes plus two fallbacks (multi-label; a cell may match
several), and writes \texttt{stats.json}, four figures, and two tables. Every number,
table, and figure in this paper is reproduced by running these two scripts in sequence;
\texttt{stats.json} is published so values can be diffed without re-running.

\subsection{Definition of the verification-failure rate}\label{sec:definition}

Because this rate is the paper's headline quantity, we state its definition explicitly
and report a denominator check. Throughout this paper,
\[
\text{verification-failed day} \equiv \text{any day whose verdict is not \verified},
\]
i.e., $\weak \cup \unverified \cup \contradicted = 227 + 108 + 19 = 354$ of $366$ days
$= 96.7\%$ (Wilson 95\% CI $[94.4\%, 98.1\%]$). We name this quantity a verification
failure rather than ``fabrication'': an \unverified{} scene may be real-but-unrecorded,
so the defensible claim is that the day \emph{failed positive scene-level verification},
not that its content is false. The definition remains deliberately conservative in the
direction of \emph{under}counting failure: machine-generated scene content must survive
positive corroboration to be counted as verified. The term \emph{fabricated} is
reserved in this paper for the \contradicted{} class only (5.2\%), whose content the
record positively disproves.

The source audit summary states its own approximate tally, ``$\approx$350 of 366 days
carry a fabricated anecdote,'' computed from pre-parse counts ($\approx$9 \verified
days) before exact table parsing. Exact parsing yields 12 \verified days. The two
figures agree within rounding of the approximate tally ($354/366$ vs.\ $\approx350/366$;
both round to $\geq$95\%). All inferential statistics in this paper use the parsed
denominators; we flag any reader-facing number that depends on which convention is used.

\subsection{Recurring false-premise screens}\label{sec:premises}

The audit identified recurring false premises --- specific false claims about the life
that resurface across many days. To make their footprint reproducible rather than
anecdotal, each premise was operationalized as a fixed regular-expression screen over the
audit rows' gist and note text (patterns fixed verbatim in
\texttt{02\_analyze.py}). These are co-occurrence screens: a day counts if its audit row
references the premise theme, not because a human re-confirmed causality per day. Counts
are therefore ceilings for per-premise incidence and may overlap across premises (three
dates appear under two premises).

\subsection{Limitations of single-rater verdicts}\label{sec:single-rater}

We describe the rating instrument exactly as it is. The 366 verdicts were assigned by an
LLM auditor working against the named independent sources: every month-table row carries
an evidence note citing specific source locations, and the rubric and precedence rule
were fixed before tallying. The author (subject) adjudicated a small number of critical
items only --- \textbf{four} explicit written confirmations are recorded in the audit
summary --- and those rulings take precedence over any other row. No second rater
participated in the original audit, so no inter-rater reliability statistic accompanied
the labels, and rater drift across the twelve month tables cannot be measured post hoc.

The consequence must be stated plainly: the dependent variable of this study is an LLM
judgment, which means this paper measures LLM confabulation \emph{using LLM-produced
labels} --- a circularity we do not pretend away. Section~\ref{sec:reliability} reports a
blind independent re-rate of a 60-day sample by a second rater from a different model
family, designed to bound exactly this concern.

\section{Results}\label{sec:results}

\subsection{The headline distribution}\label{sec:headline}

\begin{figure}[t]
\centering
\includegraphics[width=0.8\linewidth]{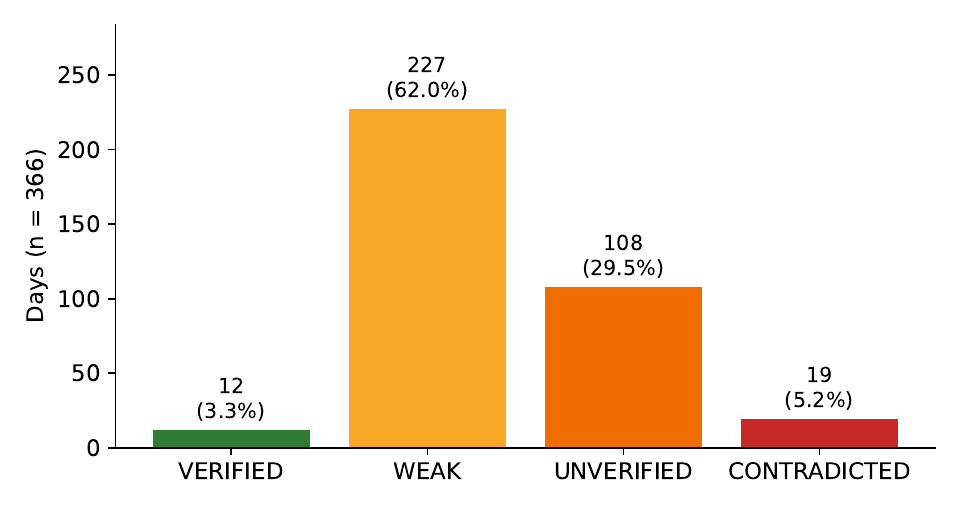}
\caption{Verdict distribution, $n=366$.}
\label{fig:verdicts}
\end{figure}

Table~\ref{tab:verdicts} and Figure~\ref{fig:verdicts} give the verdict distribution over
all 366 days. Under the definition of Section~\ref{sec:definition}, \textbf{354 of 366
days (96.7\%, Wilson 95\% CI 94.4--98.1\%) failed positive scene-level verification}.
Only 12 days
(3.3\%) contain a scene that survived positive corroboration. At the other extreme, 19
days (5.2\%, CI 3.3--8.0\%) assert claims the record actively \emph{contradicts} ---
the machine did not merely imagine, it misstated documented fact about a documented
person. In the original audit, the bulk of verification failures is \weak{} --- real settings,
employers, and people arranged inside invented scenes --- and \weak{} is the single
largest category in all three independent ratings of the sampled data, though its
measured share is rater-sensitive (Section~\ref{sec:reliability}: 43--82\% across
raters).

\begin{table}[htbp]
\centering
\caption{Scene-level verdicts over 366 generated day-narratives. ``Failed verification'' is any
non-\verified day (Section~\ref{sec:definition}); the source summary's approximate
tally ($\approx$350/366) is consistent with this exact count.}
\label{tab:verdicts}
\input{tables/verdicts.tex}
\end{table}

\subsection{Monthly variation}\label{sec:monthly}

\begin{figure}[t]
\centering
\includegraphics[width=0.95\linewidth]{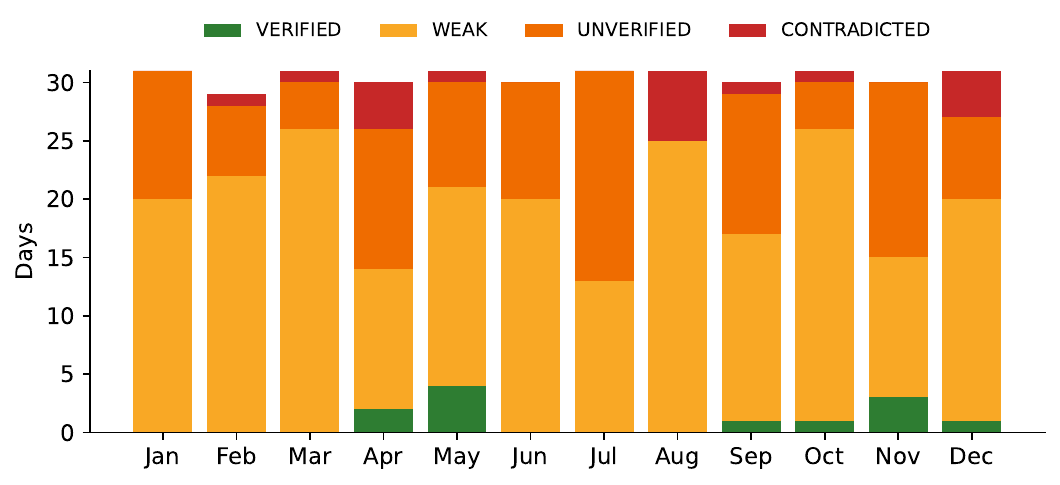}
\caption{Verdicts by month, $n=366$.}
\label{fig:monthly}
\end{figure}

Figure~\ref{fig:monthly} shows the cross-tabulation by month. Three patterns stand out.
First, verification is rare everywhere and clustered: six months (January--March, June,
July, August) contain \emph{zero} \verified days; May contributes 4 of the 12. Second,
\contradicted days are unevenly clustered rather than uniform: August (6), April (4),
and December (4) account for 14 of 19. The repository records no generation order or
dates for the month files, so we draw no chronological conclusion from this clustering.
Third, the \unverified share
swells mid-year (July: 18 of 31 days), marking stretches where entries detached from the
corpus entirely and invented generically.

\subsection{Recurring false premises (RQ2)}\label{sec:premises-results}

\begin{figure}[t]
\centering
\includegraphics[width=0.85\linewidth]{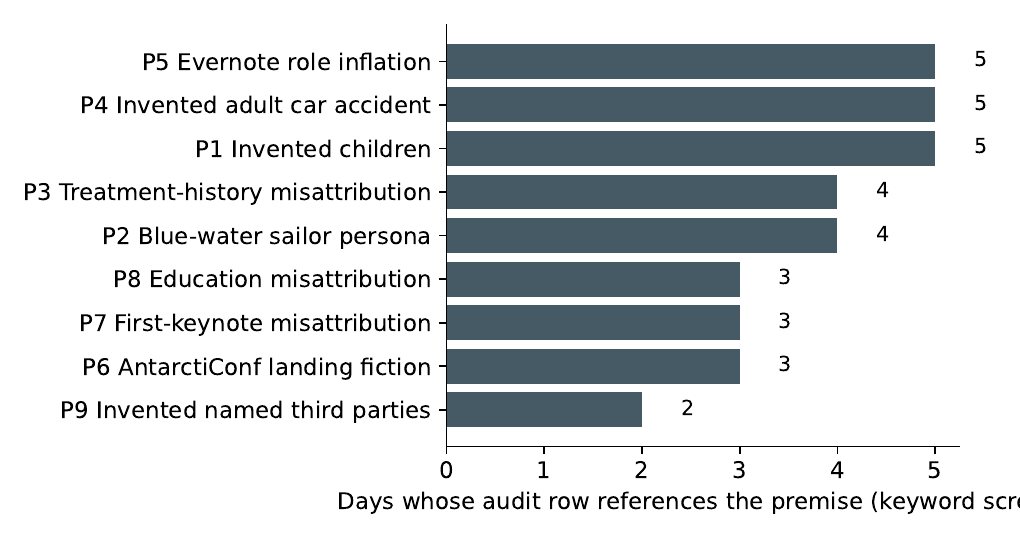}
\caption{Days whose audit row references each false premise (keyword screen).}
\label{fig:premises}
\end{figure}

Table~\ref{tab:premises} and Figure~\ref{fig:premises} show the nine recurring false
premises and their screen footprints (31 distinct days; three dates screen under two
premises). The taxonomy spans four fabrication families: \emph{invented relations}
(nonexistent children, named strangers); \emph{invented capabilities and events}
(blue-water sailing, an adult car crash, a conference keynote on Antarctica);
\emph{misattributed facts} (treatment history, first keynote, education, Evernote role);
and \emph{inverted dispositions} (fearing a diagnosis that in fact went unnoticed). One
screen (P8, education) was overturned by same-day author adjudication --- the education
content it flags was author-confirmed as real --- and is retained in the table to
document the instrument's correction trail, not as a confirmed error count.

\begin{table}[htbp]
\centering
\small
\begin{tabular}{llc}
\toprule
Premise & Family & Screen hits \\
\midrule
P1 Invented children / ``the kids'' & invented relations & 5 \\
P2 Blue-water sailor persona & invented capability & 4 \\
P3 Treatment-history misattribution & misattributed fact & 4 \\
P4 Invented adult car accident & invented event & 5 \\
P5 Evernote role inflation & misattributed fact & 5 \\
P6 AntarctiConf ``on the ice'' fiction & invented event & 3 \\
P7 First-keynote misattribution & misattributed fact & 3 \\
P8 Education misattribution & misattributed fact (reversed on adjudication) & 3 \\
P9 Invented named third parties & invented relations & 2 \\
\bottomrule
\end{tabular}
\caption{Recurring false premises; fixed keyword screens over audit rows
(Section~\ref{sec:premises}). Per-day lists: Appendix~\ref{app:premises}.}
\label{tab:premises}
\end{table}

\subsection{What the audit checked against}\label{sec:sources}

\begin{figure}[t]
\centering
\includegraphics[width=0.85\linewidth]{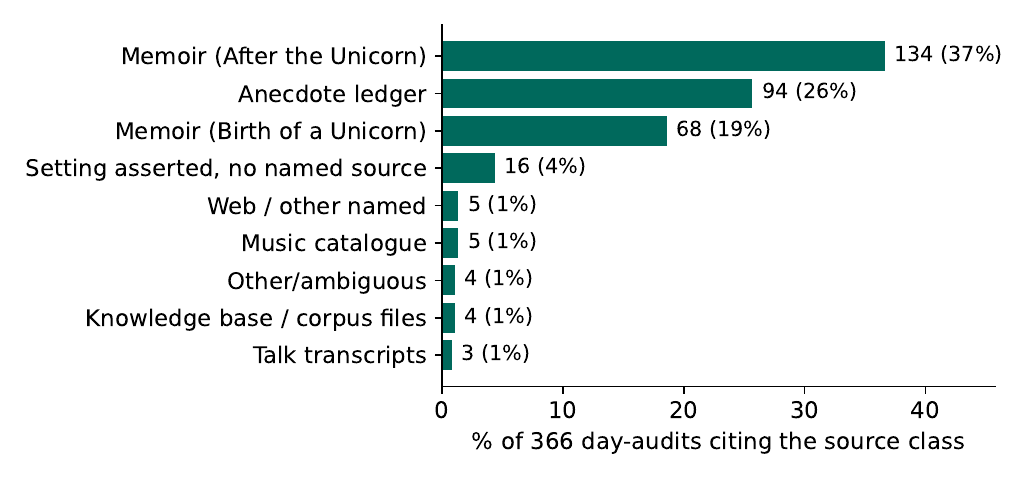}
\caption{Source-class coverage of the 252 sourced audit rows (multi-label).}
\label{fig:sources}
\end{figure}

Of 366 audit rows, 114 (31.1\%) cite \emph{no} independent source --- the day's scene had
no anchor in the record at all. The remaining 252 rows carry 333 source-class mentions
(multi-label): memoir \emph{After the Unicorn} 134 (53\% of sourced rows), anecdote
ledger 94 (37\%), memoir \emph{Birth of a Unicorn} 68 (27\%), asserted-real settings
without a named source 16, and music catalogue, web, knowledge-base, and talk-transcript
citations in single digits each (Figure~\ref{fig:sources}). The memoirs and the ledger
--- curated, author-controlled documents --- carry over 80\% of the verification load,
which is itself a finding: personal archives are the only ground truth that exists for a
life.

\subsection{What survives, and what fails}\label{sec:examples}

Table~\ref{tab:examples} lists all 12 \verified days; the survivors are almost entirely
public-career facts (Evernote tenure and scale metrics) or scenes the subject has told
publicly for years. Appendix~\ref{app:contradicted} lists all 19 \contradicted days with
their gists. The asymmetry is instructive: what survives is what the corpus says often;
what fails is what \emph{sounds like} the life --- poignant, quotable, structurally
perfect for a page-a-day format.

\begin{table}[htbp]
\centering
\small
\caption{All 12 days whose central scene was positively corroborated.}
\label{tab:examples}
\input{tables/examples.tex}
\end{table}

\subsection{Inter-rater reliability: two independent re-raters}\label{sec:reliability}

Because the original labels are themselves LLM-produced
(Section~\ref{sec:single-rater}), two independent raters --- both from a different model
family than the original auditor --- re-rated random samples of 60 of the 366 days
(uniform day-of-year draw, fixed seed 20260821, published before any rating), applying
the same verbatim rubric to the same long-form entries against the same ground-truth
corpus. Their exposure conditions differed, and we disclose both. \textbf{Rater~B}
performed a source-provenance task in the same session before rating and saw the
original audit rows for two days (neither in its sample), the audit summary's four
author confirmations, and aggregate counts; all 60 of its ratings were recorded and
frozen before any comparison. By contrast, the replication raters D and E of
Section~\ref{sec:replication} were fully blind. \textbf{Rater~C} was fully blind: a fresh session, forbidden from reading
the audit files, the derived dataset, and this paper.

\textbf{Binary result (the headline's foundation).} On whether a day's scene survives
verification --- VERIFIED vs.\ not --- raw agreement is \textbf{95.0\%} (rater~B, 57/60)
and \textbf{98.3\%} (rater~C, 59/60). We deliberately report no binary Cohen's $\kappa$:
it is degenerate here. At a VERIFIED base rate of $\approx$1--5\%, and with rater~C
assigning VERIFIED to \emph{zero} of its 60 days, one response column has no variance and
$\kappa$ collapses toward zero regardless of agreement; quoting it would understate
near-perfect raw agreement. The substantive finding is unambiguous: every independent
rating places the verification-failure rate at or above the original audit's level (B: 58/60 = 96.7\%;
C: 59/60 = 98.3\%), and rater~C found \emph{zero} corroborated scenes. The original
audit was, if anything, generous --- \textbf{the re-ratings provide no evidence that
the original verification-failure rate was inflated}.

\textbf{Four-way taxonomy.} Exact four-way agreement is 63.3\% ($\kappa=0.387$, ``fair'')
for rater~B and 81.7\% ($\kappa=0.573$, ``moderate'') for rater~C. Two observations
matter. First, the \emph{clean} rater agreed more than the partially exposed one
($0.573 > 0.387$): exposure did not inflate agreement. Second, the disagreement is not
diffuse --- it concentrates almost entirely on one boundary.

{\sloppy
\textbf{Named finding: the WEAK cell is doing two jobs.} Of rater~B's 22 disagreements,
\textbf{21 (95\%) involve WEAK} as one of the two labels: WEAK--UNVERIFIED 13
(59\%); CONTRADICTED--WEAK 5 (23\%); VERIFIED--WEAK 3 (14\%);
CONTRADICTED--UNVERIFIED 1 (5\%). The root cause is a
definitional gap in the instrument, verified by hand-arbitration of a full case.
\emph{Worked example (January 22).} The entry asserts ``At Evernote, I was working
80-hour weeks.'' Evernote employment is thoroughly documented; the literal figure appears
nowhere in the record (the memoir's only ``eighty hour'' string is ``four hundred and
eighty hours'' about an unrelated ROCeteer training event --- itself a cautionary example
of keyword-screen false positives). The original rater called this UNVERIFIED (no anchor
for the claim); both re-raters called it WEAK (real setting, unsupported specific).
\emph{Both readings are faithful to the rubric}, which defines WEAK as ``real
setting/person, invented specific scene'' and UNVERIFIED as ``no anchor'' but never
defines what counts as the \emph{setting}. Scene-level fabrication audits are thus
reliable at the binary level (does this day's scene check out?) and unreliable at the
taxonomic level for reasons internal to the rubric --- an instrument finding transferable
to any scene-level audit of generated text.\par}

\textbf{Tie-break rule (post hoc; not applied to reported labels).} To close the gap for
future use of the instrument: \emph{a day is WEAK only if a specific, named, datable
entity from the record appears inside the asserted scene; naming a real employer or city
as mere backdrop, without an episode, is UNVERIFIED.} We flag that this rule was derived
after seeing the disagreements and has not been applied to any label reported in this
paper.

\textbf{Reading.} The study's backbone --- nearly every generated day fails verification
--- replicates under blinded re-rating and strengthens with stricter blinding; the
sampled re-ratings provide no evidence that the original rate was inflated. The four-way taxonomy should be read as one rater-pair's operationalization of
a fuzzy boundary: headline percentages are binary or CONTRADICTED-based and survive
re-rating; the WEAK share carries the uncertainty quantified here.

\begin{table}[htbp]
\centering
\small
\begin{tabular}{@{}lrrrr@{}}
\toprule
\multicolumn{5}{c}{\textbf{(a) Rater B} (partial exposure)} \\
Original & V & W & U & C \\
\midrule
VERIFIED & 1 & 2 & 0 & 0 \\
WEAK & 1 & 20 & 10 & 4 \\
UNVERIFIED & 0 & 3 & 16 & 1 \\
CONTRADICTED & 0 & 1 & 0 & 1 \\
\midrule
\multicolumn{5}{c}{\textbf{(b) Rater C} (fully blind)} \\
Original & V & W & U & C \\
\midrule
VERIFIED & 0 & 1 & 0 & 0 \\
WEAK & 0 & 38 & 0 & 0 \\
UNVERIFIED & 0 & 9 & 10 & 0 \\
CONTRADICTED & 0 & 1 & 0 & 1 \\
\bottomrule
\end{tabular}
\caption{Blind re-rates of two independent 60-day samples (seed 20260821): original
labels (rows) vs.\ second raters (columns); V=VERIFIED, W=WEAK, U=UNVERIFIED,
C=CONTRADICTED. Four-way agreement: B 63.3\% ($\kappa=0.387$), C 81.7\%
($\kappa=0.573$). Binary VERIFIED-vs-not raw agreement: B 95.0\%, C 98.3\% (binary
$\kappa$ degenerate and not quoted; Section~\ref{sec:reliability}).}
\label{tab:agreement}
\end{table}

\subsection{Replication: does a 2026 model still confabulate?}\label{sec:replication}

The original generator is now dated --- July--August 2025 (Section~\ref{sec:limitations})
--- so a referee may ask whether the measurement is obsolete. We therefore re-ran the
generation under controlled conditions on the same 60-day sample, with a current,
named model and the \emph{same documented inputs recovered for the 2025 run}: the
verbatim archived user
prompt and template (exemplar days included), plus only the day's quote from
\texttt{commonsense.csv}. Three arms were run. \textbf{Arm~B}: \texttt{openai/gpt-5.4}
(same vendor as the original pipeline), ungrounded. \textbf{Arm~B2}:
\texttt{anthropic/claude-sonnet-5}, ungrounded (cross-vendor replicate). \textbf{Arm~C}:
\texttt{openai/gpt-5.4} grounded --- given verbatim excerpts retrieved from the
subject's ground-truth corpus for its day's quote and instructed to use only
corroborated material and to say when it does not know; i.e., the paper's own
remediation recommendation implemented cheaply. The 60 \emph{original} entries
constitute Arm~A1.

\textbf{Blind rating, and a disclosed failure of blindness.} All 240 entries (A1, B,
B2, C) were anonymized into shuffled sets with opaque identifiers and rated against the
same four-verdict rubric by two independent raters in fresh sessions --- rater~D for
arms B/B2/C (180 items), rater~E for A1 (60 items) --- each seeing only entry text,
rubric, and excerpts retrieved from the entry's text alone; the orchestrator verified
that neither rater accessed the arm map, prior ratings, results files, arm folders, or
this paper. We report a first rating pass that \emph{failed this standard}: it was
performed by the same agent that generated the entries while holding the arm map, and
was therefore structurally compromised. It was discarded and re-run independently.
Agreement between the compromised pass and the independent blind pass was 74.4\% exact
($\kappa=0.383$, $n=180$): the headline conclusions were unchanged, but per-item labels
diverged substantially, so the non-blind pass should be relied on for nothing finer
than the binary rate. This $\kappa$ sits alongside 0.387 and 0.573 measured earlier on
the same rubric (Section~\ref{sec:reliability}) --- repeated evidence that the taxonomy
is rater-sensitive while the binary judgment is robust.

\textbf{Results} (Table~\ref{tab:replication}). Three findings, stated at the strength
the data support. \textbf{(1) No improvement was observed under the reconstructed documented-input
condition}: both 2026
models fail verification in 100\% of sampled days when given the 2025 inputs, statistically
indistinguishable from the 2025 originals (Fisher exact, A1 vs.\ B $p=1.0$; A1 vs.\ B2
$p=1.0$). The recency objection to this paper is answered with data.
\textbf{(2) Not vendor-specific}: OpenAI and Anthropic entries fail verification for
every one of the 120 combined ungrounded days (zero corroborated scenes; B vs.\ B2
$p=1.0$).
\textbf{(3) Grounding is the only intervention that moved the measure}: the
verification-failure rate drops from 100\% to 83.3\% under corpus grounding (B vs.\ C $p=0.0013$; B2 vs.\ C
$p=0.0013$), corroborated scenes rise from 0 to 10 (A1 vs.\ C $p=0.0084$) --- direct
evidence for the remediation recommendation, though even grounding leaves five of six
days below the verification bar.

\begin{table}[htbp]
\centering
\small
\caption{2025-vs-2026 replication on the same 60-day sample (seed 20260821); final
fully-blind ratings (raters D/E). ``Failed verification'' = any non-\verified{} day.}
\label{tab:replication}
\resizebox{\linewidth}{!}{\input{tables/replication.tex}}
\end{table}

\textbf{Confound disclosure.} Arms differ in vendor \emph{and} release date, so no
clean model-generation effect is claimed beyond the $p=1.0$ nulls above; A1 was rated
in a separate batch by a different rater than B/B2/C, so its comparison is between
raters as well as arms; and the 2025 artifact came from ~10 days of iterative
human-guided chat plus a Cursor agent session while B/B2/C are single-shot API calls
--- the interaction pattern differs, not just the model. The raters also operate from
retrieved excerpts rather than the full corpus and assigned zero \contradicted{}
verdicts in any arm, so cross-arm comparisons are valid while absolute VERIFIED counts
are instrument-specific.

\section{Discussion}\label{sec:discussion}

\textbf{Grounded drift, not unanchored invention.} The dominant failure class is
\weak{} --- real setting, invented scene --- and it is the single largest category in
all three independent ratings (43\%, 63\%, and 82\% of sampled days). We report the
spread rather than a point estimate because the WEAK/UNVERIFIED boundary is the least
reliable distinction in the instrument (Section~\ref{sec:reliability});
notably, the two re-raters erred in \emph{opposite} directions relative to the original
(one shifted WEAK days toward UNVERIFIED, the other toward WEAK), so the spread reflects
a definitional boundary, not correctable systematic bias. The qualitative claim is what
is stable, and it is the one surface-fact evaluations miss: the generator knew the real employers,
the real city, the real boat --- and then invented the scene around them, fluently and
in the first person. This matches the confabulation account of \citet{sui2024confabulation}
and explains why attitude-level fidelity in person simulators \citep{park2024selfreports}
coexists with our result: a system can match survey responses while producing
unsupported scenes for nearly every episodic narration it gives.

\textbf{Verification failure is structural, not sporadic.} With 96.7\% of days failing
verification, no sampling-of-errors account is tenable: days with corroborated scenes were the
exception, not corrections of a normally reliable process. (Archived prompts that could
explain the mechanism were not kept, so we make no prompt-level claim.) The
monthly view sharpens this: zero verified days for six months of the year, and
contradictions concentrated in specific months rather than spread uniformly. The
nine false premises show a further regularity --- the same specific falsehoods (the
children, the ocean crossing, the adult car crash) recur across months, behaving like
attractors of the generation process rather than independent coin flips.

\textbf{Contradiction is the tip that reveals the iceberg.} Only 19 days are provably
false, but their existence certifies that \unverified days cannot be read as ``probably
fine.'' When the verification record positively contradicts 5.2\% of generated scenes,
absence of evidence elsewhere is weak reassurance. This ordering ---
positive corroboration required, silence insufficient --- is exactly the audit stance we
recommend for any persona system deployed about a real person.

\textbf{Remediation.} The same project produced a workflow that survives contact with
these numbers. First, \emph{author adjudication takes precedence}: machine-verifiable
facts are checked against sources, but critical personal facts get explicit written
rulings from the subject, which bind all downstream edits (this rule corrected one audit
ruling the same day it was issued). Second, \emph{lesson-only rewriting}: days with no
corroborated story are rewritten to carry the reflection without the invented scene,
after expanding a bank of verified anecdotes mined from the broader corpus. Both moves
are cheap precisely because the audit instrument localizes which days need them.

\textbf{For digital twins specifically.} The subject maintains a curated knowledge base
for a professional digital twin in the same ecosystem --- but that knowledge base was
never supplied to the generator (Section~\ref{sec:limitations}), so no amount of twin
fidelity could have kept this autobiography honest: only scene-level auditing did.
Fidelity claims for
person simulations should therefore state which layer they cover: profile, attitudes, or
episodes. The episodic layer, our data suggest, is where trust goes to die.

\section{Limitations}\label{sec:limitations}

\textbf{Single subject.} $n=366$ days is one life, one corpus, one generator deployment.
The measurement is deep but not a sample across people; nothing here estimates a
population fabrication rate. The design's justification is the opposite of breadth: no
laboratory can easily assemble a comparably documented ground truth for an arbitrary
subject, and the single-subject audit is exactly the instrument an individual can reuse
on their own twin.

\textbf{LLM-assigned labels (circularity).} As Section~\ref{sec:single-rater} details,
the original verdicts were assigned by an LLM auditor; only four items carry explicit
author rulings. The paper therefore measures LLM confabulation with LLM-produced labels.
The rubric and precedence rule were fixed before tallying and every row preserves an
evidence note, but the labels were not blind-rated in the original audit.
Section~\ref{sec:reliability}'s blind re-rates bound this concern empirically: the
binary headline replicates under two independent raters (raw agreement 95.0\% and
98.3\%) and strengthens with stricter blinding, but the four-way taxonomy achieves only
fair-to-moderate agreement ($\kappa=0.387$ and $0.573$), with instability concentrated
in the WEAK cell; we treat class-level percentages below the VERIFIED/non-VERIFIED split
as rater-sensitive.

\textbf{LLM re-raters are not ground truth.} The blind raters of
Sections~\ref{sec:reliability} and~\ref{sec:replication} establish inter-model
\emph{consistency}, not correctness: agreement between models could coexist with shared
error, and near-zero \verified{} prevalence makes high binary agreement easy to attain.
The obvious next step --- a blinded human adjudication sample, which this study's
unusual design makes feasible, since the author is the subject --- is flagged as future
work rather than claimed here.

\textbf{Finite ground truth cuts both ways.} \verified is a strict lower bound on true
accuracy: a scene absent from the memoirs, ledger, and corpus is unproven, not false.
Consequently \unverified (29.5\%) conflates ``invented'' with ``real but unrecorded.''
We mitigate this by (i) naming the headline measure a \emph{verification-failure
rate} rather than a fabrication rate --- failure to verify is not proof of falsehood
--- and (ii) reserving \emph{fabricated}/false for the positively disproven
\contradicted class. The headline should be read with exactly this asymmetry in mind: 96.7\% of days
fail verification; 5.2\% are provably false; the remainder are unproven.

\textbf{Generator identity: now documented, with residual gaps.} Session-log recovery
(PROVENANCE analysis; \texttt{analysis/05\_provenance.py}) identified the generator of
record for the 366 entries: OpenAI \textbf{o3-pro} designed the daily-entry template
(``Daily Inspiration Template,'' 31 messages, July~7--17, 2025) and \textbf{gpt-4o},
\textbf{o3}, and \textbf{o4-mini-high} generated entries from the quote list
(``Daily Entry Template,'' January~1--10 verbatim from the same prompt archived with
this paper; ``Bio-based Anecdote Generation'' continued to October~2025), with a
Cursor composer run (August~1, 2025) executing a one-shot version of the same prompt.
The per-day split of authorship among these models is not recoverable, the Cursor-side
model is not recorded in local stores, and no log of the July~13, 2026 audit session or
the page-a-day compression exists on this machine (sweeps documented in
\texttt{results/provenance.json}), so those producers remain UNKNOWN. We therefore
cannot attribute failure rates to a single model version --- the original measurement
characterizes a documented three-model ChatGPT pipeline under iterative human guidance,
and Section~\ref{sec:replication}'s replication provides the controlled single-model
comparison that the historical artifact cannot.

\textbf{Scene-level granularity.} The audit grades the day's central anecdote scene, not
every atomic claim \citep{min2023factscore}. A \weak day may contain individually true
sentences; the unit of analysis is the memory a reader would take away.

\textbf{Keyword screens.} Premise footprints are co-occurrence screens over gist and note
text, fixed for reproducibility but not human-verified per day; counts are ceilings, and
one screen (P8) was reversed by adjudication and is retained only as a correction-trail
record.

\textbf{Recency of the historical measurement.} The audited entries were generated
July--August 2025, so the headline 96.7\% characterizes that generation's models.
Section~\ref{sec:replication} addresses this directly: current models from two vendors,
given identical inputs, fail verification in 100\% of sampled days ungrounded --- the finding
has aged \emph{into} the present, not out of it. The residual confound (iterative
human-guided chat in 2025 vs.\ single-shot calls in 2026) is disclosed there and
prevents a pure model-effect claim in either direction.

\textbf{Remediation is designed, not yet evaluated at scale.} The lesson-only rewrite
and adjudication workflow were specified during this project; their effect on reader
trust is unmeasured. The grounding arm of Section~\ref{sec:replication} evaluates the
retrieval half of the workflow at the generation stage under fully-blind rating and
finds a significant verification-failure reduction (100\%$\to$83.3\%, $p=0.0013$; corroborated
scenes 0$\to$10), with substantial residual failure.

\subsection{Use of generative AI tools}\label{sec:aiuse}

Per arXiv policy on significant use of generative AI tools, we disclose plainly how this
paper --- and its underlying dataset --- were produced. No AI system is an author; the
human author takes full responsibility for all contents.

\textbf{The audit labels are an LLM product.} The 366 scene-level verdicts were
assigned by an LLM auditor (Claude-based agents, July~13, 2026) working against named
independent sources and writing a per-row evidence note citing specific source
locations; they were not assigned by a human. The human author's role in label
production was adjudication of critical items only: four explicit written rulings
(counted from the audit files), binding under the summary's written precedence rule.
The label set predates this paper by five weeks and was not generated at any point for
it --- that remains true and important, but it does not make the labels human work.
The dependent variable of this study is therefore an LLM judgment; that circularity is
stated in Sections~\ref{sec:single-rater} and~\ref{sec:limitations} and is why the
blind independent re-rates of Section~\ref{sec:reliability} exist.

\textbf{This paper's own production.} The prose draft, analysis scripts
(\path{analysis/01_parse_factcheck.py}, \path{analysis/02_analyze.py}), and
literature-search assistance were drafted by an LLM (``ox-alpha,'' an anonymous stealth
model served via OpenRouter, August 2026) working autonomously from the author's own
pre-existing data and under her direction, with a second LLM (Claude) orchestrating the
workflow and independently verifying its output. Verification actually performed on the
AI-drafted material: every headline number was independently re-derived from the
dataset before drafting (now published as date+verdict only); all 17 citations were verified against Crossref,
arXiv, and OpenAlex APIs with the query trail logged; and one earlier version of this
disclosure falsely described the audit labels as exclusively human work --- caught in
external review and corrected here. A paper whose subject is machine-generated
fabrication treats this disclosure, including its corrections, as part of its evidence
base.

\section{Ethics}\label{sec:ethics}

\textbf{Consent and dual role.} The audited subject and the paper's author are the same
person, who commissioned the audit of her own records, adjudicated its critical items,
and publishes this paper about it; consent is therefore structural rather than
procedural. No human-subjects data
were collected: every analyzed document pre-existed the study.

\textbf{Privacy.} The paper reports machine-generated content about one individual with
her participation, and no content from files marked sensitive in the source repository
was read or used. Health information appears only as required to document the audit's
author-confirmed rulings (specifically, that generated treatment-history claims were
false); no diagnostic detail beyond those rulings is included. Third-party named
individuals appear only where they already appear in the subject's own published corpus
and tables --- public-record professional associations --- plus two \emph{invented}
names (premise P9), which are flagged as fabrications appearing in no real source and
are attributable to no real person.

\textbf{Self-audit risk.} A single author auditing her own life could bias labels toward
a preferred narrative. The mitigations are structural: the rubric predates tallying,
every verdict carries a recorded evidence note and source citation, critical items carry
written rulings that bind the summary, and the full month tables remain available for
inspection. We disclose the residual risk rather than claim it away.

\section{Reproducibility Statement}\label{sec:repro}

\paragraph{Replication package.} Code, derived data, and all computed results are public at
\url{https://github.com/heathriel/synthetic-memoir-audit}. The package contains the analysis
scripts, \path{stats.json}, the blind-replication and inter-rater agreement results, the recovered
provenance data, the LaTeX tables, the four figures, the verified bibliography, and a 366-row
verdict dataset (\path{results/factcheck_verdicts_redacted.csv}: date and verdict for every day).

The package \emph{deliberately omits} the 366 source day-entries, the 240 replication entries, and
the per-day anecdote-gist, evidence-note and source columns of the audit dataset. The audited corpus
consists of machine-generated first-person narratives about a private individual, a large majority of
which fail verification; publishing it would distribute those claims about a named person in
perpetuity. The headline distribution, confidence intervals, and monthly cross-tabs depend only on date and verdict, which are published for all 366 days. Premise screens, source-class coverage, and example tables depend on withheld audit fields; their computed outputs are published but cannot be independently regenerated from the public package. Raw entries are
available from the author on request.

The public entry point is \texttt{analysis/02\_analyze\_public.py}
(stdlib only), which runs from the published verdict dataset and reproduces exactly:
$n=366$; VERIFIED 12 / WEAK 227 / UNVERIFIED 108 / CONTRADICTED 19;
verification-failure 96.7\% (CI 94.4--98.1); contradicted 5.2\% (CI 3.3--8.0);
and the monthly cross-tab. The published verdict dataset reproduces the headline
distribution, monthly cross-tabs, and associated confidence intervals. Premise
screens, source-class coverage, and example tables depend on withheld audit fields;
their computed outputs and the parser integrity report are published, but they cannot
be regenerated from the public package alone. The upstream parser
\path{analysis/01_parse_factcheck.py} --- which converts the twelve audit month
tables into per-day rows (366 rows, zero duplicate keys, zero unparsed verdicts) ---
requires the unpublished audit corpus and therefore cannot be run from the package.
The provenance recovery
(\texttt{analysis/05\_provenance.py} $\to$ \texttt{provenance.json}) and the
2025-vs-2026 replication comparison (\texttt{07\_blind\_replication.py} $\to$
\path{blind_replication_results.json}) are likewise published at the level of
results, not raw material: the generated entry texts, rating inputs and transcripts,
and arm maps are withheld under the privacy rule above. Replication generators were
\texttt{openai/gpt-5.4} and \texttt{anthropic/claude-sonnet-5}; final ratings are the
fully-blind independent raters' (raters D/E; Section~\ref{sec:replication}); all API
calls ran on August~22, 2026, with token usage published
(\path{replication_usage.json}). The rubric and summary quoted in
Section~\ref{sec:taxonomy} are dated July~13, 2026, five weeks before analysis. All 17
bibliography entries were verified against arXiv, Crossref, or OpenAlex APIs on
August~21, 2026, with the query trail logged (\texttt{RESEARCH\_LOG.md}); no unverified
reference is cited. Figures regenerate via published code (Python~3, matplotlib).

\paragraph{Venue note.} The draft uses the plain \texttt{article} class for arXiv
(cs.AI primary, with cs.CL and cs.CY as cross-lists) submission. For venue submission,
swap the preamble to the target class
(e.g., \texttt{acmart} for FAccT/CSCW lanes, ACL template for hallucination-workshop
lanes): content requires no changes; tables use \texttt{booktabs} and figures are
standalone PDFs.

\appendix

\section{All Contradicted Days}\label{app:contradicted}

Table~\ref{tab:contradicted} lists every day rated \contradicted, with abbreviated gists
from the audit rows (trims mark elided list tails).

\begin{table}[h]
\centering
\small
\begin{tabular}{ll>{\raggedright\arraybackslash}p{8.6cm}}
\toprule
Date & Premise & Anecdote gist (abbreviated) \\
\midrule
February 7 & P6 & AntarctiConf ``application'' sat in drafts; scared to submit; ``I submitted. I got in.'' \\
March 26 & --- & ``Months before diagnosis I lived in fear of cancer, imagined it in my body'' \\
April 4 & P4 & Hospital bed after a car hit her, unsure she'd walk again; married Matt 2019 \\
April 14 & P4 & Thanksgiving overeating; ``after the car accident,'' healing, pushed too hard \\
April 15 & P6/P7 & First keynote ON Antarctica, ``standing on the seventh continent,'' mic might freeze \\
April 19 & P4 & Decade of trying to be liked; then ``a car hit me,'' hospital bed, came out sharper \\
May 4 & --- & ``Learn Spanish'' sat on to-do list 3 yrs, never attempted; Evernote thousands$\to$millions \\
August 2 & --- & ``Deserve trap''; sailboat in Mexico ``supposed to be my reward\ldots I'd earned it'' \\
August 8 & P1 & Reading calendar after diagnosis; ``RV trip to Disney with the kids'' \\
August 10 & --- & 50th birthday, Matt surprise party; AntarctiConf ``researchers who'd lost colleagues\ldots'' \\
August 17 & P1 & ``If it fits perfectly, inflate standards''; Phil Libin quote; Evernote ``inflation pressure'' \\
August 24 & P6 & Non-negotiables; ``AntarctiConf, water was life\ldots every drop purified'' \\
August 29 & P1/P6 & Location history as priorities; ``after the RV trip with the kids'' \\
September 2 & P3 & ``Chemotherapy took the choice away''; forced rest during chemo births best ideas \\
October 2 & P6 & Night before AntarctiConf, female ``co-organizer'' finds her at 2am \\
December 2 & P3 & Cancer ``found during a routine checkup''; body kept score of skipped meals \\
December 3 & P2 & ``Preparing to sail to New Zealand,'' fielding everyone's advice \\
December 11 & P2 & ``I started sailing across oceans and the weather didn't care'' \\
December 23 & P2 & ``Boat started taking on water --- I took the wheel,'' years of drilled procedures \\
\bottomrule
\end{tabular}
\caption{All 19 \contradicted days. The Premise column reports keyword-screen output over
gist+note text (Section~\ref{sec:premises}); themes need not be visible in the
abbreviated gist. ``---''
marks screen-negative days flagged during audit rather than by keyword screen.}
\label{tab:contradicted}
\end{table}

\section{Replication Methods}\label{app:replication-methods}

This appendix makes Section~\ref{sec:replication} auditable from the paper alone;
the published scripts and result files back every item (raw generation outputs and
rating transcripts are withheld under the privacy rule of
Section~\ref{sec:repro}).

\textbf{Sampling.} 60 of 366 days, drawn once with
\path{random.Random(20260821).sample(range(1,367),60)} over a leap-year day-of-year
population; the mapping and day list are published
(\path{results/revalidation_sample.json}). Per month: Jan 5, Feb 8, Mar 2, Apr 4,
May 5, Jun 6, Jul 4, Aug 6, Sep 4, Oct 5, Nov 6, Dec 5.

\textbf{Generation inputs.} Each call received (i) the verbatim archived user prompt
--- instruction, entry template, and two exemplar days carrying personal facts
(\path{results/original_generation_prompt.txt}, published) --- followed by (ii) the fixed
continuation: ``Using the commonsense.csv -- please generate the entry for this quote,
using the template above: [Month Day] ``quote'' (mood/theme: mood).'' Nothing else was
supplied to Arms B/B2. Arm~C received the same plus (iii) retrieved excerpts (below)
and the grounding instruction to use only corroborated material and to say when it
does not know, and not to invent employers, events, people, or dates.

\textbf{Models and API settings.} Via the OpenRouter API on August~22, 2026:
Arms B and C used \texttt{openai/gpt-5.4}; Arm~B2 used
\texttt{anthropic/claude-sonnet-5}; temperature 0.7; \texttt{max\_tokens} 1400;
per-call token usage is published (\path{replication_usage.json}).

\textbf{Retrieval (Arm C).} Keywords were extracted from the day's quote
(\texttt{[a-zA-Z']\{4,\}}, stopword-filtered, first six); every line of the
ground-truth corpus (HEATHER\_CORPUS, 19 memoir chapters, anecdote ledger, Birth of a
Unicorn text) was scored by keyword count, and the top six unique three-line windows
were supplied verbatim with file/line locators. The identical procedure, run on each
entry's own text, supplied excerpts to the raters.

\textbf{Rating.} Each of the 240 entries was placed in a shuffled, anonymized file
consisting of the verbatim rater prompt below, the retrieved excerpts, and the entry
text; raters saw nothing else. Rater~D rated arms B/B2/C (180 items); rater~E rated A1
(60 items); both worked in fresh sessions, and access logs were verified to show zero
reads of the arm map, prior ratings, results files, arm folders, or the paper. Rater
prompt, verbatim: ``You are auditing a first-person diary entry written for a real
person (Heather Wilde) by an AI system. Rate ONLY the central anecdotal scene (the
personal story the entry is built around) against this four-verdict rubric:
VERIFIED --- the specific scene is corroborated by the supplied source excerpts; WEAK
--- a real setting/person/employer appears, but the specific scene/quote/detail is
unsupported by the excerpts; UNVERIFIED --- no anchor in any excerpt; generic
invention; CONTRADICTED --- an excerpt positively disproves a stated detail. Judge
only from the excerpts provided. If the entry asserts nothing checkable, use
UNVERIFIED. Respond with STRICT JSON only: \{verdict: VERIFIED|WEAK|UNVERIFIED|
CONTRADICTED, evidence: $\leq$40 words\}.'' The raters' own session models are not
recorded in this repository; the join/unblind script
(\texttt{analysis/07\_blind\_replication.py}) is published, while the raters' verdict
transcripts are withheld under the privacy rule of Section~\ref{sec:repro}.

\textbf{Failure handling.} API calls retried up to four times with increasing backoff;
a generation was accepted only as a complete response (no partial outputs), and all
180 generations ultimately succeeded. The first rating pass (performed non-blind by
the generating agent) was discarded wholesale, not merged; its divergence is reported
in Section~\ref{sec:replication}.

\section{Premise Screen Day Lists}\label{app:premises}

Fixed-screen hits per premise (Section~\ref{sec:premises}):\
P1: Mar~13, Mar~16, Aug~8, Aug~17, Aug~29.\
P2: Oct~10, Dec~3, Dec~5, Dec~23.\
P3: Sep~2, Oct~3, Oct~17, Oct~30.\
P4: Apr~4, Apr~14, Apr~19, May~30, Oct~3.\
P5: Jun~10, Jul~3, Jul~7, Aug~3, Aug~20.\
P6: Apr~15, Aug~29, Oct~6.\
P7: Jan~29, Apr~8, Apr~15.\
P8 (reversed on adjudication): Apr~25, Aug~14, Oct~16.\
P9: Jul~13, Jul~29.

\bibliographystyle{unsrtnat}
\bibliography{references}

\end{document}

%% file: tables/verdicts.tex
\begin{tabular}{lrr}
\toprule
Verdict & Days & \% of 366 \\
\midrule
VERIFIED & 12 & 3.3\% \\
WEAK & 227 & 62.0\% \\
UNVERIFIED & 108 & 29.5\% \\
CONTRADICTED / flagged & 19 & 5.2\% \\
\midrule
Failed verification (any non-VERIFIED class) & 354 & 96.7\% \\
\bottomrule
\end{tabular}

%% file: tables/examples.tex
\begin{tabular}{ll>{\raggedright\arraybackslash}p{7.2cm}}
\toprule
Date & Verdict & Anecdote gist (abbreviated) \\
\midrule
April 2 & VERIFIED & Joined Evernote as employee number eight; a mentor said "build a culture" \\
April 6 & VERIFIED & Hit one hundred million users at Evernote; no celebration \\
May 9 & VERIFIED & Lived on a sailboat in Mexico \& an RV at Disney World; travel light \\
May 13 & VERIFIED & Asked every founder "what aren't you saying?"; never turned it on her own dying marriage\ldots \\
May 14 & VERIFIED & Underestimated as a woman at THQ; winning College Jeopardy sealed the lesson \\
May 30 & VERIFIED & Near-death car accident as a teen (distracted driver); fear rode along for years \\
September 25 & VERIFIED & "Overnight success in daily installments" --- Evernote 100M users, employee \#8 \\
October 4 & VERIFIED & Earns Phil Libin's trust "mile by mile" as Evernote employee \#8; parallels dating Matt\ldots \\
November 20 & VERIFIED & Built the opposite of chasing a unicorn valuation --- ROCeteer as a community model\ldots \\
November 21 & VERIFIED & Ciara, documentary filmmaker, "accidental school-garden program" in Las Vegas $\to$ largest\ldots \\
November 26 & VERIFIED & College interview at Kodak: "how many gas stations in the US?"; interviewer: "I don't know" \\
December 5 & VERIFIED & Downsizing a house to a sailboat forced ruthless minimalism \\
\bottomrule
\end{tabular}

%% file: tables/replication.tex
\begin{tabular}{lrrrrrr}
\toprule
Arm & VERIFIED & WEAK & UNVERIFIED & CONTRADICTED & Verification failure & 95\% CI \\
\midrule
A1: 2025 (orig.\ audit) & 1 & 16 & 43 & 0 & 98.3\% & [91.1, 99.7] \\
B: gpt-5.4 & 0 & 1 & 59 & 0 & 100.0\% & [94.0, 100.0] \\
B2: cl.-sonnet-5 & 0 & 14 & 46 & 0 & 100.0\% & [94.0, 100.0] \\
C: gpt-5.4 grounded & 10 & 26 & 24 & 0 & 83.3\% & [72.0, 90.7] \\
\bottomrule
\end{tabular}

%% file: references.bib
@article{huang2025survey,
  author  = {Huang, Lei and Yu, Weijiang and Ma, Weitao and Zhong, Weihong and Feng, Zhangyin and Wang, Haotian and Chen, Qianglong and Peng, Weihua and Feng, Xiaocheng and Qin, Bing and Liu, Ting},
  title   = {A Survey on Hallucination in Large Language Models: Principles, Taxonomy, Challenges, and Open Questions},
  journal = {ACM Transactions on Information Systems},
  year    = {2025},
  doi     = {10.1145/3703155},
  note    = {arXiv:2311.05232}
}

@article{ji2023survey,
  author  = {Ji, Ziwei and Lee, Nayeon and Frieske, Rita and Yu, Tiezheng and Su, Dan and Xu, Yan and Ishii, Etsuko and Bang, Ye Jin and Madotto, Andrea and Fung, Pascale},
  title   = {Survey of Hallucination in Natural Language Generation},
  journal = {ACM Computing Surveys},
  volume  = {55},
  number  = {12},
  pages   = {248:1--248:38},
  year    = {2023},
  doi     = {10.1145/3571730}
}

@article{sui2024confabulation,
  author  = {Sui, Peiqi and Duede, Eamon and Wu, Sophie and So, Richard Jean},
  title   = {Confabulation: The Surprising Value of Large Language Model Hallucinations},
  journal = {arXiv preprint arXiv:2406.04175},
  year    = {2024}
}

@article{zhang2023siren,
  author  = {Zhang, Yue and Li, Yafu and Cui, Leyang and Cai, Deng and Liu, Lemao and Fu, Tingchen and Huang, Xinting and Zhao, Enbo and Zhang, Yu and Xu, Chen and Chen, Yulong and Wang, Longyue and Luu, Anh Tuan and Bi, Wei and Shi, Freda and Shi, Shuming},
  title   = {Siren's Song in the {AI} Ocean: A Survey on Hallucination in Large Language Models},
  journal = {arXiv preprint arXiv:2309.01219},
  year    = {2023}
}

@article{tonmoy2024mitigation,
  author  = {Tonmoy, S. M Towhidul Islam and Zaman, S M Mehedi and Jain, Vinija and Rani, Anku and Rawte, Vipula and Chadha, Aman and Das, Amitava},
  title   = {A Comprehensive Survey of Hallucination Mitigation Techniques in Large Language Models},
  journal = {arXiv preprint arXiv:2401.01313},
  year    = {2024}
}

@article{alansari2026comprehensive,
  author  = {Alansari, Aisha and Luqman, Hamzah},
  title   = {Large Language Models Hallucination: A Comprehensive Survey},
  journal = {Computer Science Review},
  year    = {2026},
  doi     = {10.1016/j.cosrev.2026.100970},
  note    = {arXiv:2510.06265}
}

@inproceedings{min2023factscore,
  author    = {Min, Sewon and Krishna, Kalpesh and Lyu, Xinxi and Lewis, Mike and Yih, Wen-tau and Koh, Pang Wei and Iyyer, Mohit and Zettlemoyer, Luke and Hajishirzi, Hannaneh},
  title     = {{FActScore}: Fine-grained Atomic Evaluation of Factual Precision in Long Form Text Generation},
  booktitle = {Proceedings of the 2023 Conference on Empirical Methods in Natural Language Processing (EMNLP)},
  year      = {2023},
  note      = {arXiv:2305.14251}
}

@article{wei2024longform,
  author  = {Wei, Jerry and Yang, Chengrun and Song, Xinying and Lu, Yifeng and Hu, Nathan and Huang, Jie and Tran, Dustin and Peng, Daiyi and Liu, Ruibo and Huang, Da and Du, Cosmo and Le, Quoc V.},
  title   = {Long-form Factuality in Large Language Models},
  journal = {arXiv preprint arXiv:2403.18802},
  year    = {2024}
}

@article{wang2025refact,
  author  = {Wang, Yindong and Prei{\ss}, Martin and Bugue{\~n}o, Margarita and Hoffbauer, Jan Vincent and Ghajar, Abdullatif and Buz, Tolga and de Melo, Gerard},
  title   = {{ReFACT}: A Benchmark for Scientific Confabulation Detection with Positional Error Annotations},
  journal = {arXiv preprint arXiv:2509.25868},
  year    = {2025}
}

@article{park2023generative,
  author  = {Park, Joon Sung and O'Brien, Joseph C. and Cai, Carrie J. and Morris, Meredith Ringel and Liang, Percy and Bernstein, Michael S.},
  title   = {Generative Agents: Interactive Simulacra of Human Behavior},
  journal = {arXiv preprint arXiv:2304.03442},
  year    = {2023},
  note    = {UIST 2023}
}

@article{park2024selfreports,
  author  = {Park, Joon Sung and Zou, Carolyn Q. and Kamphorst, Jonne and Egan, Niles and Shaw, Aaron and Hill, Benjamin Mako and Cai, Carrie and Morris, Meredith Ringel and Liang, Percy and Willer, Robb and Bernstein, Michael S.},
  title   = {{LLM} Agents Grounded in Self-Reports Enable General-Purpose Simulation of Individuals},
  journal = {arXiv preprint arXiv:2411.10109},
  year    = {2024},
  note    = {v1 titled ``Generative Agent Simulations of 1,000 People''}
}

@article{wang2025beyondprofile,
  author  = {Wang, Zixiao and Zhang, Duzhen and Agrawal, Ishita and Gao, Shen and Song, Le and Chen, Xiuying},
  title   = {Beyond Profile: From Surface-Level Facts to Deep Persona Simulation in {LLMs}},
  journal = {arXiv preprint arXiv:2502.12988},
  year    = {2025}
}

@article{xie2024psydt,
  author  = {Xie, Haojie and Chen, Yirong and Xing, Xiaofen and Lin, Jingkai and Xu, Xiangmin},
  title   = {{PsyDT}: Using {LLMs} to Construct the Digital Twin of Psychological Counselor with Personalized Counseling Style for Psychological Counseling},
  journal = {arXiv preprint arXiv:2412.13660},
  year    = {2024}
}

@article{chan2024falsememories,
  author  = {Chan, Samantha and Pataranutaporn, Pat and Suri, Aditya and Zulfikar, Wazeer and Maes, Pattie and Loftus, Elizabeth F.},
  title   = {Conversational {AI} Powered by Large Language Models Amplifies False Memories in Witness Interviews},
  journal = {arXiv preprint arXiv:2408.04681},
  year    = {2024}
}

@article{tang2024rolebreak,
  author  = {Tang, Yihong and Wang, Bo and Wang, Xu and Zhao, Dongming and Liu, Jing and Zhang, Jijun and He, Ruifang and Hou, Yuexian},
  title   = {{RoleBreak}: Character Hallucination as a Jailbreak Attack in Role-Playing Systems},
  journal = {arXiv preprint arXiv:2409.16727},
  year    = {2024}
}

@article{kong2024sharp,
  author  = {Kong, Chuyi and Luo, Ziyang and Lin, Hongzhan and Fan, Zhiyuan and Fan, Yaxin and Sun, Yuxi and Ma, Jing},
  title   = {{SHARP}: Unlocking Interactive Hallucination via Stance Transfer in Role-Playing {LLMs}},
  journal = {arXiv preprint arXiv:2411.07965},
  year    = {2024}
}

@article{akpinar2025whosasking,
  author  = {Akpinar, Nil-Jana and Lee, Chia-Jung and Murdock, Vanessa and Perona, Pietro},
  title   = {Who's Asking? Evaluating {LLM} Robustness to Inquiry Personas in Factual Question Answering},
  journal = {arXiv preprint arXiv:2510.12925},
  year    = {2025}
}
